\documentclass[twocolumn,english]{article}
\usepackage[T1]{fontenc}
\usepackage[latin9]{inputenc}
\usepackage{geometry}
\usepackage{amsmath}
\usepackage{graphicx}

\makeatletter
\@ifundefined{date}{}{\date{}}
\usepackage{booktabs}
\usepackage{url}
\usepackage{indentfirst}

\author{
{\large\textbf{Roberto Rey-de-Castro}}\\
{\large\textbf{Princeton University}}\\
rrey@princeton.edu\\
  \and
    {\large\textbf{Herschel Rabitz}}\\
{\large\textbf{Princeton University}}\\
hrabitz@princeton.edu\\
}

\makeatother

\usepackage{babel}
\begin{document}

\title{\textbf{Targeted Nonlinear Adversarial Perturbations in Images and
Videos}}
\maketitle
\begin{abstract}
We introduce a method for learning adversarial perturbations targeted
to individual images or videos. The learned perturbations are found
to be sparse while at the same time containing a high level of feature
detail. Thus, the extracted perturbations allow a form of object or
action recognition and provide insights into what features the studied
deep neural network models consider important when reaching their
classification decisions. From an adversarial point of view, the sparse
perturbations successfully confused the models into misclassifying,
although the perturbed samples still belonged to the same original
class by visual examination. This is discussed in terms of a prospective
data augmentation scheme. The sparse yet high-quality perturbations
may also be leveraged for image or video compression.
\end{abstract}

\section{Introduction\label{sec:Introduction}}

A longstanding goal in the field of artificial intelligence (AI) is
the realization of artificial general intelligence (AGI) \cite{goertzel2007artificial}
which would enable a computer to perform any intellectual task accessible
to a human brain. One of the many challenges to be overcome before
achieving AGI is the development of computer models with enough capacity
to process the highly complex phenomena customarily managed by a human
brain. Another outstanding challenge is finding practical ways for
comprehensively conveying to algorithms what a human understands from
each data sample. While full AGI has proven elusive, there has been
considerable recent progress in enabling machines to emulate humans
in specialized tasks. Notably, advances in image classification \cite{krizhevsky2012imagenet,DBLP:journals/corr/SimonyanZ14a,he2016deep},
language translation \cite{cho2014learning}, self-driving car technology
\cite{bojarski2016end}, cancer detection \cite{cruz2013deep}, etc.
often achieve near- or sometimes super-human performance. This recent
progress has been enabled by the application of deep neural network
architectures \cite{lecun2015deep} that allow a machine to learn
complex abstract representations. A question relevant to both AGI
and specialized AI is to what extent current algorithms capture the
relevant representations or features that enable human reasoning.
Answering this question requires, as a first step, translating into
human terms what a computer model has considered essential in reaching
its decisions \cite{gunning2017explainable}. One method for extracting
this type of information consists of introducing perturbations into
the inputs of a computer model and then inspecting the model responses
\cite{ribeiro2016should,fong2017interpretable,szegedy2013intriguing,goodfellow6572explaining}.
Relevant perturbations that obscure features important to the model
will have a large effect on the model outputs. Here we report on a
nonlinear perturbation approach capable of learning optimal perturbations
that drastically change the original model output while keeping the
input as close as possible to the original one. The perturbation method
is applied to image and video classification models. Importantly,
the perturbations are generated by deep convolutional neural networks
(convnets) able to learn complex (yet minimalistic) perturbations
targeted to each particular sample. In implementing the technique
we confirmed the results of previous research in that while the studied
models do consider some features that are indeed relevant to humans
\cite{zeiler2014visualizing,zhou2014object,fong2017interpretable,ribeiro2016should},
the models are also easily confused by the introduction of seemingly
subtle perturbations \cite{szegedy2013intriguing,goodfellow6572explaining,kurakin2016adversarial,liu2016delving,baluja2017adversarial,moosavi2017universal}
that do not change the image or video classification from a human's
perspective. To mitigate the latter issue we consider enhancements
to data augmentation procedures as suggested by Szegedy et. al. \cite{szegedy2013intriguing}.
The process of data augmentation can be greatly improved through the
addition of perturbations known to drastically affect the performance
of a model (to be improved through re-training) but that do not change
the classification of an object from the point of view of a human
assessment \cite{szegedy2013intriguing,goodfellow6572explaining,madry2017towards}.
Enhancing data augmentation through perturbations provides with a
way of ``telling'' the model (to be re-trained) what a human understands
but was not possible to convey to it during the first round of model
training. Furthermore, since the learned perturbations tend to expose
the most salient features of an object or action in question, the
learned perturbations can be applied to perform object or action detection
in images or videos. This extracted information also opens a window
into what the model considered important in reaching its decisions
in terms understandable by humans. Finally, the observed sparsity
of the most relevant perturbations may be leveraged for a novel form
of image and video compression and approximate reconstruction.

Our specific contributions may be summarized as follows:
\begin{enumerate}
\item We introduce a method for learning adversarial perturbations targeted
to a specific sample image or video.
\item We introduce a form of object or action detection/explanation of model
outcomes based on the outstanding perturbations.
\item We propose an iterative model training procedure relying on image
or video data augmentation to be repeated until a measurable score
reaches certain values.
\item We propose a method for image or video compression and approximate
reconstruction.
\end{enumerate}

\section{Related work\label{sec:Related-work}}

\subsection{Perturbations of images}

Soon after the release of the first widely successful deep learning
models for image classification, perturbation schemes were proposed
for understanding them \cite{zeiler2014visualizing} or for exposing
some of their shortcomings through the generation of adversarial perturbations
that confuse the models into misclassifying \cite{szegedy2013intriguing}.
These two areas of study using perturbation methods have seen considerable
progress over the years.

Following the seminal work of Szegedy et. al. \cite{szegedy2013intriguing}
on adversarial perturbations, the linear nature of convolutional layers
(before activation functions) was exploited for the rapid generation
of adversarial examples \cite{goodfellow6572explaining}. The existence
of (untainted) real-world images that also cause deep learning models
to misclassify has also been reported \cite{kurakin2016adversarial}.
These and other findings triggered a wave of introspection resulting
on a series of comprehensive studies such as an evaluation of the
robustness of neural networks \cite{carlini2016towards,arpit2017closer},
model generalization assessments \cite{zhang2016understanding,arpit2017closer},
and a proof of the existence of universal adversarial perturbations
that can be applied to any image \cite{moosavi2017universal}. An
impressive consequence of the latter result may be found in the work
of Baluja and Fisher \cite{baluja2017adversarial} who were able to
train adversarial models on a full dataset of images with the explicit
goal of confusing a target model into believing that all of the samples
belong to an arbitrarily chosen class.

The intrinsic complexity of deep neural networks has led many researchers
into attempting to understand model decisions through perturbative
tests. Understanding the outcomes of deep learning models has attracted
much attention due to its potential for accelerating the field's progress
and rate of adoption \cite{gunning2017explainable} and due to recent
changes in regulatory requirements \cite{albrecht2016gdpr}. Perturbational
approaches for understanding the outcomes of image classification
models include greedy search approaches in which parts of the image
are occluded \cite{zeiler2014visualizing} or grayed out \cite{zhou2014object}
until it is misclassified. In these cases the result is a saliency
map that indicates how strongly the pixels in the image correlate
with the output class score. Thus, the most intense regions in each
map are considered to be related to image features important in the
corresponding model decision. It is also possible to construct a simpler
local model around a given image (in the space of inputs) by sampling
around it through the introduction of small perturbations \cite{ribeiro2016should}.
The sampled points are often fitted by a linear regressor with L1
regularization (e.g., Lasso) which enables an identification of the
image's most salient features from the model's perspective. Finer
image features that are relevant to the model's decision can be exposed
by iteratively optimizing the values of a multiplicative gray scale
mask \cite{fong2017interpretable}. The mask values determine the
degree of blurring in the perturbed image and noise is introduced
to mitigate optimization artifacts. 

To different degrees, most of the perturbation schemes mentioned above
are able to perform a form of object localization by extracting a
mask (often defined as the difference between the perturbed and original
images) with most intense values around the object in question.

\subsection{Perturbations of videos}

The only previous work that we are aware of regarding perturbations
of video samples for the study of deep neural network models was recently
published by Wei et. al. \cite{wei2018sparse}. In this study, video
pixels are modified to reduce the class score assigned to the original
video while attempting to keep the perturbed video as close as possible
to the original one. Modifications to the original video are confined
to a few video frames by the use in the temporal direction of a \emph{L1}
norm penalty during the optimization and by the introduction of a
mask that explicitly prevents modifications to some frames. The authors
were able to produce sparse adversarial perturbations that considerably
reduced the original score assigned by the model.

\subsection{Perturbative data augmentation}

Due to the large number of parameters present in deep neural networks,
generally a very large number of training samples is necessary to
prevent overfitting. This fact was recognized early on in the development
of modern convnet architectures \cite{krizhevsky2012imagenet} and
prompted the development of data augmentation techniques. Early image
data augmentation was achieved by shifting, rotating, cropping, and
flipping the images \cite{krizhevsky2012imagenet,howard2013some}.
More recently, other heuristics have been used for data augmentation
such as color casting, vignetting and lens distortion \cite{wu2015deep}
leading to the training of more robust image classification models.
In general, any transformation that does not change the class of a
sample can be exploited for data augmentation. Thus, adversarial image
perturbations generated by a learned perturbative mask \cite{szegedy2013intriguing,goodfellow6572explaining,madry2017towards}
or through the training of generative adversarial networks (GANs)
\cite{antoniou2017data} have been proposed for data augmentation.
GANs have been successfully applied to the augmentation of biomedical
images \cite{calimeri2017biomedical} field in which the number of
available samples is limited.

\subsection{Video and image reconstruction}

Deep neural networks of various architectures have been applied to
image and video reconstruction tasks. These include image denoising
\cite{agostinelli2013adaptive,burger2012image,schuler2013machine,sun2015learning,vincent2010stacked,xu2014deep},
image super-resolution \cite{cui2014deep,dong2016image}, video super-resolution
\cite{huang2015bidirectional}, and video compressive sensing \cite{wang2015lisens,iliadis2018deep}.

\section{Learned perturbations\label{sec:Learned-perturbations}}

A procedure for learning minimalistic input perturbations that drastically
modify the outputs of image or video classification models is described
next. Assume that there is a trained model $M$ that transforms an
input array $\textbf{X}$ into an output array $\textbf{\textbf{y}}$,
i.e., 
\[
\textbf{y}=M(\textbf{X})
\]
The elements of output array $\textbf{y}$ indicate the probabilities
of associated classes. Let us denote by $k_{1},\ldots,k_{5}$ the
indices corresponding to the largest 5 output values $y_{k_{1}},\ldots,y_{k_{5}}$.
The goal here is to obtain a perturbed input $\textbf{X}'$ with associated
output $\textbf{y}'=M(\textbf{X}')$ such that:
\begin{enumerate}
\item $\sum_{i=1}^{5}\left|y'_{k_{i}}\right|\ll1$ (i.e., the perturbed
scores for the originally top 5 classes are minimized). 
\item $\left\Vert \textbf{X}-\textbf{X}'\right\Vert \ll1$ (i.e., the perturbed
input is as close as possible to the original input).
\end{enumerate}
\begin{figure}
\centering

\includegraphics[scale=0.25]{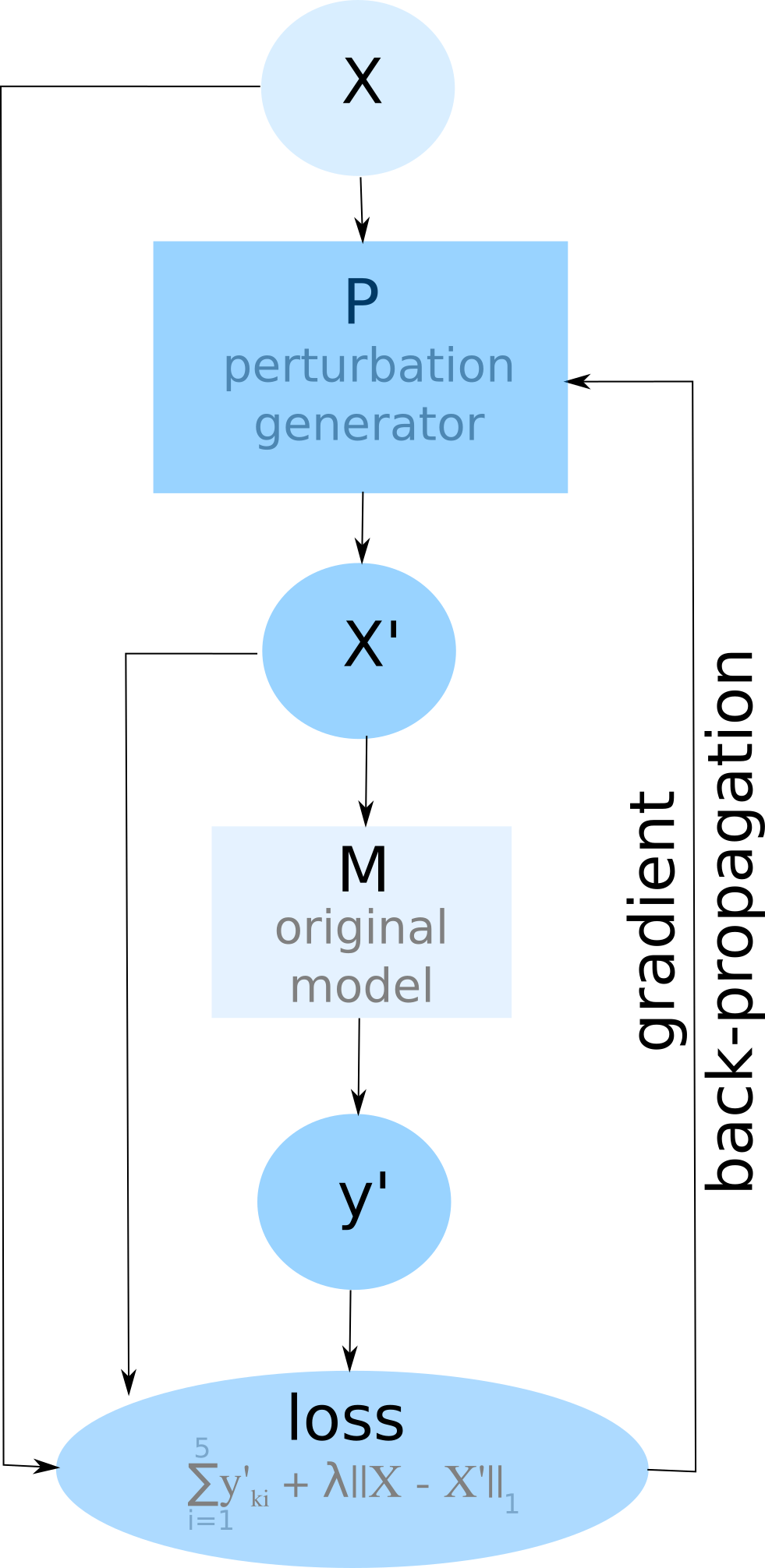}


\caption{\label{fig:Block-diagram-pert}Block diagram illustration of the optimum
perturbation search procedure. The input $\textbf{x}$ and the model
$M$ are shaded to indicate that their values or parameters are kept
constant during loss minimization.}
\end{figure}
In order to accomplish both aims above another model $P$ is used
to produce the perturbed inputs $\textbf{X'}=P(\textbf{X})$ and the
parameters of $P$ are iteratively modified to approach both conditions.
Figure \ref{fig:Block-diagram-pert} illustrates the adaptive procedure
used to optimize the parameters of model $P$. In mathematical form,
the optimum parameters of model $P$ are the approximate result of
minimizing a loss function expressing the two conditions above, i.e., 

\begin{equation}
\underset{P}{\text{min}}\left[\sum_{i=1}^{5}M\left(P(\textbf{X})_{k_{i}}\right)+\lambda\left\Vert \textbf{X}-P(\textbf{X})\right\Vert _{1}\right]\label{eq:loss_min_top5}
\end{equation}
where $\left\Vert \textbf{X}\right\Vert _{1}$ denotes the \emph{L1}
norm of tensor $\textbf{X}$ and $\lambda$ is a constant for balancing
the two terms in the loss. By minimizing the top 5 original outputs,
as opposed to only a single top class, the algorithm above ensures
that all the important features present in the input are considered
(e.g., otherwise small perturbations could be inserted to simply transform
a dog image from one breed of dog to another similar one). This is
particularly important in cases when the top class score is not outstanding.

It is also possible to attempt maximizing an specified class with
index $m$, in this case the parameters of model $P$ are modified
such that
\begin{equation}
\underset{P}{\text{min}}\left[\sum_{i=1}^{5}\left(1-M(P(\textbf{X})_{m}\right)+\lambda\left\Vert \textbf{X}-P(\textbf{X})\right\Vert _{1}\right]\label{eq:loss_max_m}
\end{equation}

Section \ref{sec:Experiments} shows the results of implementations
of the perturbation procedure to image and video models.

\section{Experiments\label{sec:Experiments}}

\subsection{Implementation details}

There is wide versatility in the choice of the perturbation generator
$P$. In general, a suitable choice for $P$ is any model architecture
able to produce meaningful perturbations on the input. In the present
case $P$ was taken as a convnet for generating perturbations on both
images and videos. The particular convnet architecture used here consisted
of three blocks each containing three convolutional layers. Each block
was followed by a rectified linear (relu) layer. All convolutional
layers had 3 channels (corresponding to the three basic colors) and
their kernels were taken as (3,3) or (3,3,3) for the analysis of image
or video models, respectively. The convolutional layer padding was
chosen so as to preserve the original input dimensions. The loss given
by eq. (\ref{eq:loss_min_top5}) or eq. (\ref{eq:loss_max_m}) was
iteratively reduced by a gradient descent algorithm (Adam optimizer
\cite{DBLP:journals/corr/KingmaB14}) that modified all the convolutional
weights (filters) and biases of the perturbation generator $P$. The
convolutional layer weights and biases were initialized such that
tensors passed through them unmodified during the first epoch.

The core of the code was written using the pytorch deep learning framework
\cite{paszke2017automatic}. The analyzed image classification model
(VGG19 \cite{DBLP:journals/corr/SimonyanZ14a}) along with its pre-trained
weights were automatically downloaded through pytorch's wrapper to
the torchvision package \cite{marcel2010torchvision}. The studied
video classification model (I3D \cite{carreira2017quo}) with its
pre-trained weights were downloaded from a github repository \cite{i3d_implementation}.

Image and video manipulations made use of the scikit-image and scikit-video
packages, respectively. Image samples were downloaded from google
images. Video samples corresponding to randomly drawn rows from the
kinetics dataset \cite{kay2017kinetics} were downloaded from their
corresponding YouTube urls.

The experiments were run on a NVIDIA Titan X GPU. All the codes and
data mentioned in this paper are publicly available on github \cite{roberto1648github}.

\subsection{Perturbating images (VGG19)}

\begin{figure}
\resizebox{\columnwidth}{!}{%
\centering

\includegraphics[scale=0.3]{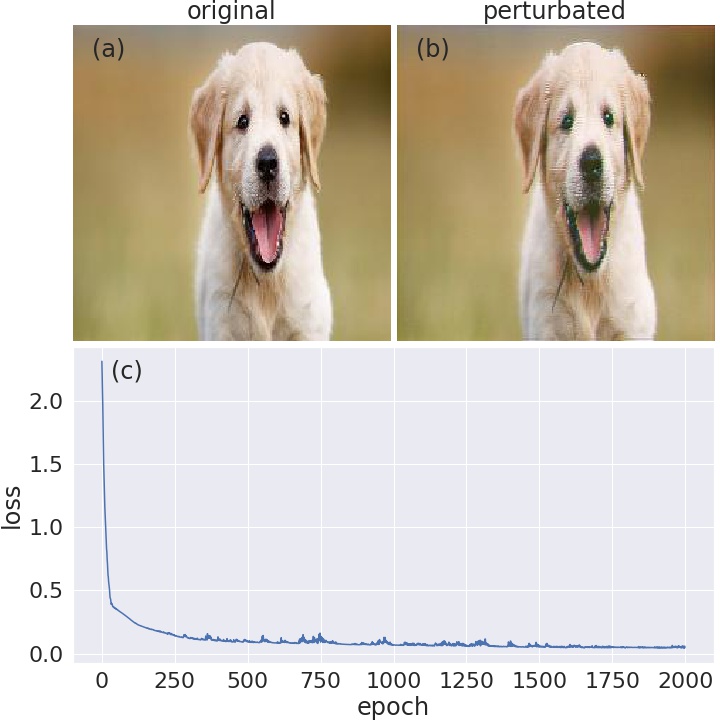}

}

\caption{\label{fig:img-loss-pert}Results of the perturbative procedure applied
to the image classification model VGG19. (a) original image sample.
(b) The optimum perturbed image that minimized the loss by simultaneously
minimizing the top 5 VGG19 scores in the original image while attempting
to keep the perturbed image as close as possible to the original one.
(c) Loss history through the learning iterations (epochs). }
\end{figure}

The broadly used VGG19 model for image classification \cite{DBLP:journals/corr/SimonyanZ14a}
was chosen as a first target model for the perturbative procedure
described in Sec. \ref{sec:Learned-perturbations}. Figure \ref{fig:img-loss-pert}
shows the results on one of the considered image samples. The original
sample (fig. \ref{fig:img-loss-pert}(a)) was first resized to 224x224
and scaled as required for the inputs of VGG19 \cite{DBLP:journals/corr/SimonyanZ14a}
and then fed to the iterative algorithm depicted in fig. \ref{fig:Block-diagram-pert}.
The loss function in eq. \ref{eq:loss_min_top5} was then minimized.
Parts (b) and (c) of fig. \ref{fig:img-loss-pert} show the optimum
perturbed image that minimized the loss and the loss history through
the learning iterations (epochs), respectively. The dog in the perturbed
image has a slightly lower contrast than in the original one. Nevertheless,
both images are very different from VGG19's point of view. The top
5 scoring image classes, as evaluated by VGG19, for the original and
perturbed images are given as ``sample 1'' entries in table \ref{tab:img-origs-perts}.
The relatively mild inserted perturbations turned the dominant VGG19
class from ``kuvasz'' (a dog breed) to ``African grey'' (a parrot).

\begin{figure}
\resizebox{\columnwidth}{!}{%
\centering

\includegraphics[scale=0.45]{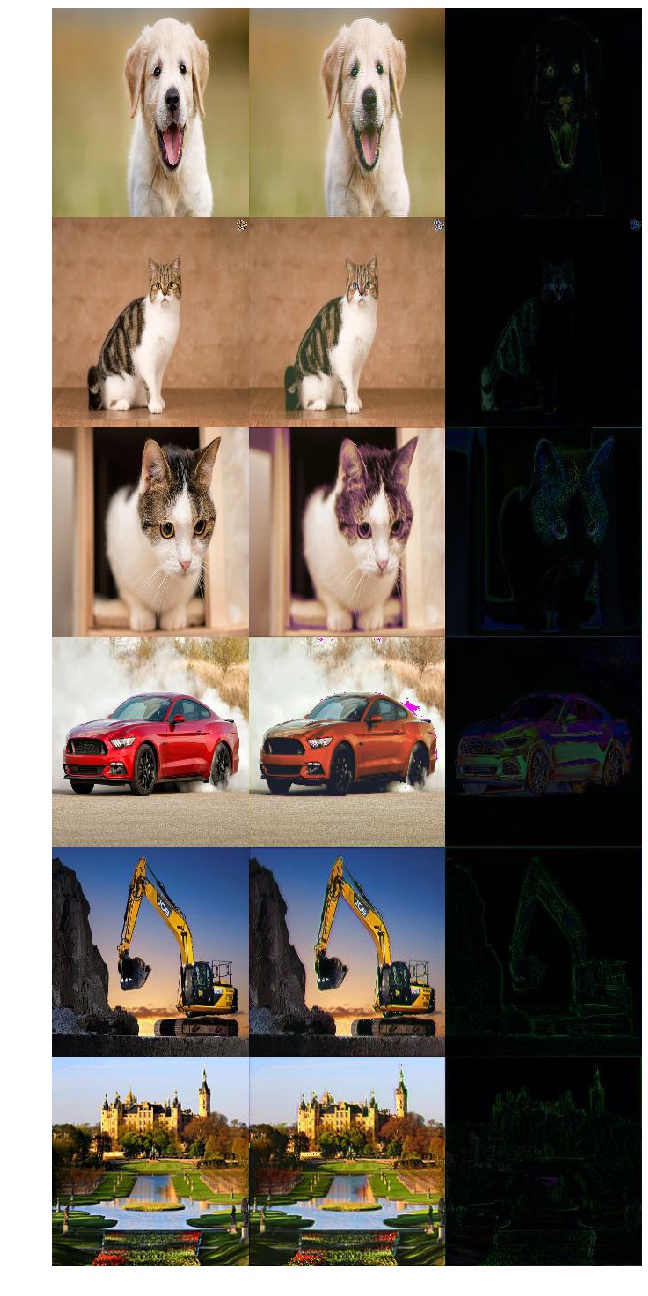}

}

\caption{\label{fig:origs-perts-diffs}(Color online) Original sample images
(first column) along with their corresponding optimally perturbed
images (second column). The absolute value of the differences between
original and perturbed images are shown on the third column. For all
cases, the VGG19 scores corresponding to the dominant classes in the
original images are drastically reduced in the perturbed ones. All
the images available on github \cite{roberto1648github}.}
\end{figure}

Optimal learned perturbations were extracted from a total of six samples
downloaded from google images. The results are shown in fig. \ref{fig:origs-perts-diffs}.
The first, second, and third columns of images correspond to the original,
perturbed, and the difference between original and perturbed images,
respectively. Since the jpeg image format only accepts image data
values in the range from 0 to 1, the absolute value of the difference
between the original and perturbed images were taken for constructing
the difference images. In all cases the VGG19 scores for the originally
dominant classes were drastically reduced. Nevertheless, the perturbed
images are remarkably similar to the original ones. The main modifications
mostly involved a slight decoloration of the target object in the
various perturbed images. This fact is reflected in the difference
images that are mostly black (i.e., zero values) but for a few colored
pixels that sketch important features of the target objects. Notice,
particularly, the cat image on the second row. In this case the perturbations
focused on the cat's features (e.g., its ears and tail) but also on
a small cartoon on the upper right corner. It turns out that the cartoon
contained a rough and undersampled sketch of a cat's face which was
also detected by the perturbation procedure. In the case of the sports
car (4-th row in the figure) the main perturbation strategy seems
to have been changing its color from bright red to a flatter orange
hue. In all the image differences, important features (sometimes of
specific colors) of the objects in question are highlighted from the
black background.

The top 5 classes along with their associated VGG19 scores for the
six image samples considered are shown in table \ref{tab:img-origs-perts}
for the original and perturbed images in fig. \ref{fig:origs-perts-diffs}.
The sample number in table \ref{tab:img-origs-perts} corresponds
to the row number (from the top to the bottom) in fig. \ref{fig:origs-perts-diffs}.
The relatively mild changes introduced into the original images resulted
in drastic reductions of VGG19 scores for the originally dominant
classes. The dominant classes for the perturbed images were in most
cases qualitatively far from the original ones. An exception was sample
5 (``sport car'') which has a dominant perturbed class given by
``amphibious vehicle'' which is somewhat related to the original
class. But even in this case the perturbed image contains practically
the same object from the point of view of a human observer.

\begin{table}
\resizebox{2.7in}{!}{%
\centering

\begin{tabular}{llllll} \toprule   & {} &              original &         &          perturbated &         \\   & {} &                 class &   score &                class &   score \\ sample & {} &                       &         &                      &         \\ \midrule 1 & {} &                kuvasz &  0.8234 &         African grey &  0.9961 \\   & {} &        Great Pyrenees &  0.1412 &               muzzle &  0.0007 \\   & {} &      golden retriever &  0.0236 &               kuvasz &  0.0003 \\   & {} &    Labrador retriever &  0.0031 &         carriage dog &  0.0003 \\   & {} &  Old English sheepdog &  0.0016 &                 swab &  0.0002 \\   & {} &                       &         &                      &         \\ 2 & {} &             tabby cat &  0.8022 &         king penguin &  0.9877 \\   & {} &             tiger cat &  0.1246 &      prairie chicken &   0.009 \\   & {} &          Egyptian cat &  0.0701 &                 hare &  0.0004 \\   & {} &           Persian cat &  0.0011 &         Egyptian cat &  0.0004 \\   & {} &               doormat &  0.0004 &            albatross &  0.0003 \\   & {} &                       &         &                      &         \\ 3 & {} &          Egyptian cat &  0.5098 &               screen &  0.3684 \\   & {} &             tabby cat &  0.2082 &              monitor &  0.3245 \\   & {} &             tiger cat &  0.0819 &           television &  0.1768 \\   & {} &         window screen &  0.0631 &    computer keyboard &  0.0433 \\   & {} &          window shade &  0.0317 &             notebook &  0.0202 \\   & {} &                       &         &                      &         \\ 4 & {} &             sport car &  0.5307 &   amphibious vehicle &  0.9783 \\   & {} &             car wheel &  0.1315 &           racing car &  0.0067 \\   & {} &                grille &  0.1093 &            speedboat &  0.0062 \\   & {} &           convertible &  0.1037 &            sport car &  0.0026 \\   & {} &            estate car &  0.0631 &            car wheel &  0.0023 \\   & {} &                       &         &                      &         \\ 5 & {} &                 crane &  0.9656 &            mousetrap &  0.2222 \\   & {} &                 wreck &  0.0204 &             cassette &  0.1466 \\   & {} &             harvester &  0.0024 &                 iPod &  0.0958 \\   & {} &        container ship &  0.0021 &     pencil sharpener &  0.0585 \\   & {} &                  plow &  0.0008 &           television &  0.0562 \\   & {} &                       &         &                      &         \\ 6 & {} &                castle &  0.7823 &                 desk &   0.415 \\   & {} &              lakeside &   0.037 &     desktop computer &  0.1414 \\   & {} &               gondola &  0.0344 &              monitor &  0.0753 \\   & {} &             submarine &   0.029 &               screen &  0.0506 \\   & {} &                  dock &  0.0265 &           restaurant &  0.0198 \\   & {} &                       &         &                      &         \\ \bottomrule \end{tabular}

}

\caption{\label{tab:img-origs-perts}Top 5 classes and associated VGG19 scores
for the original and perturbed images for the six image samples considered.}
\end{table}

\begin{figure}
\resizebox{\columnwidth}{!}{%
\centering

\includegraphics[scale=0.3]{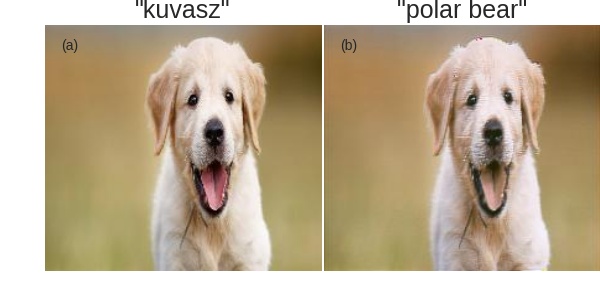}

}

\caption{\label{fig:dog-to-bear}Original (a) and perturbed (b) images corresponding
to a perturbation learned using eq. (\ref{eq:loss_max_m}) that maximized
the score for the ``polar bear'' class while attempting to keep
the perturbed image as close as possible to the original one.}
\end{figure}

As mentioned in sec. \ref{sec:Learned-perturbations}, in addition
to generating perturbations that minimize the top classes, it is also
possible to attempt a maximization of a different class. For this,
sample 1 was perturbed using eq. (\ref{eq:loss_max_m}) and setting
index $m$ to the class corresponding to ``polar bear''. The original
dog image along with the resulting optimally perturbed one are shown
in fig. \ref{fig:dog-to-bear}. As for the samples shown above, here
the perturbed image is very close to the original one while the VGG19
scores are also drastically different. It is necessary to point out
that, with the currently utilized perturbation generator, it was not
possible to maximize every arbitrary class on a given image. Gradient
descent runs on the same dog image that attempted to maximize the
scores of ``tabby cat'' or ``sea lion'' did not converge.

\subsection{Perturbating videos (I3D)}

\begin{figure}
\resizebox{\columnwidth}{!}{%
\centering

\includegraphics[scale=0.3]{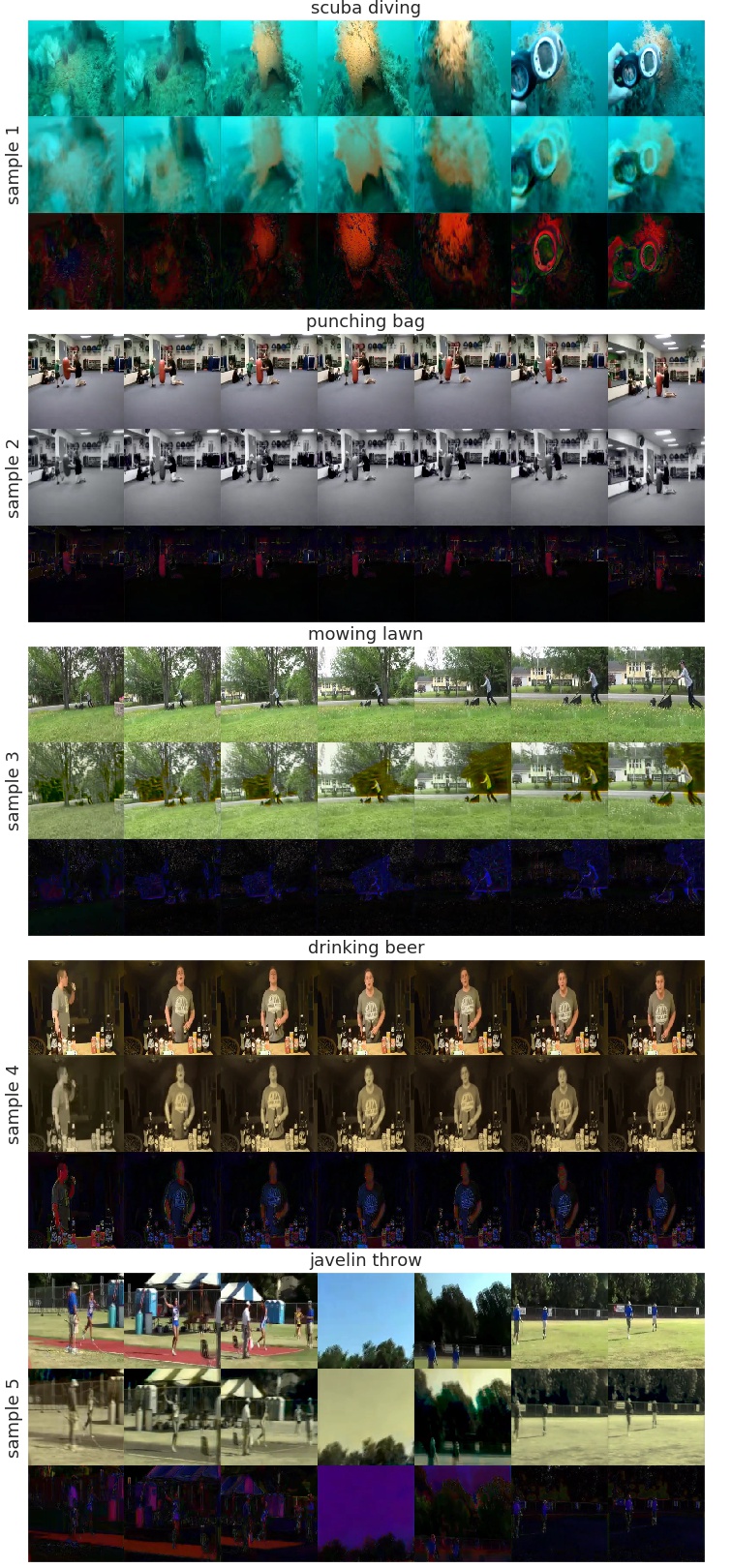}

}

\caption{\label{fig:videos}Video samples considered for perturbation analysis.
The sample number and corresponding action (i.e., the class on the
kinetics video dataset) are indicated on the right-hand side and top
of each sample, respectively. For each sample seven frames from the
original, perturbed, and difference videos are shown on the first,
second and third row, respectively. All the videos are available on
github \cite{roberto1648github}.}
\end{figure}

The video classification model I3D \cite{carreira2017quo} was also
analyzed by the perturbative procedure. Five video samples were randomly
drawn from the kinetics video dataset \cite{kay2017kinetics}. A few
frames extracted from each sample video are shown in fig. \ref{fig:videos}.
The kinetics dataset extracts 10 seconds of video from publicly available
YouTube samples. Each video is then assigned one of 400 categories
corresponding to the predominant action in the sampled portion of
the video. To perform the perturbative algorithm on the available
GPU memory, the selected videos were undersampled by keeping only
one of every 4 consecutive frames. I3D was able to correctly identify
the assigned label in the undersampled videos. The top 5 original
video classes along with their corresponding I3D scores are shown
in table \ref{tab:vid-origs-perts}. 

As was done for the case of images discussed above, the video samples
were perturbed following the algorithm in fig. \ref{fig:Block-diagram-pert}.
In this case the perturbation generator $P$ consisted of blocks of
3D convolutional layers. The optimally perturbed videos successfully
minimized the I3D scores for the top 5 original classes while keeping
the perturbed video as close as possible to the original one. Frames
from the perturbed videos are shown on the second rows (of each sample
group) in fig. \ref{fig:videos}. Whereas the perturbed videos exhibit
some blurred or decolored views, they still convey information about
the originally labeled action from the point of view of a human observer.
The third row (of each sample group) in fig. \ref{fig:videos} shows
frames from videos representing the (absolute value) differences between
the original and perturbed videos. In general, the differences have
near-zero pixel values everywhere but on pixels conveying important
aspects of the main action being undertaken in the original video.
The difference videos constitute sparse versions of the original videos
that convey to a human observer the main points in the original videos.

The strategies followed by the perturbation algorithm in reducing
the originally dominant class scores varied from sample to sample
(see fig. \ref{fig:videos}). In sample 1 the perturbation strategy
appears to have been a blurring of the images along with the removal
of the red content in some of the frames. The red extracted content
appears to come from the light shone on the coral and the scuba diver's
mask. The latter objects are informative of the main action that labels
the original video (``scuba diving'', see table \ref{tab:vid-origs-perts}).
The amount of introduced perturbations to the original frames in sample
1 is reduced by removing intensity from the red channels instead of
the predominant green channels. The blurring in the perturbed sample
1 video may be attributed to mixing in the time dimension (i.e., mixing
the pixels of adjacent frames). The perturbed video is classified
as ``smoking hookah''. While a human observer may not agree with
such a classification of the perturbed video, the blurring in the
video does make the frames images appear closer to containing smoke.
The perturbations in sample 2 (``punching bag'') removed the red
color from the punching bag which is then highlighted thorough the
difference video. The perturbations also focused on the contours of
certain objects and people in the video. The perturbed video for sample
2 is classified as ``breakdancing''. Since punching bags are often
red, the model may not have recognized the decolored bag and then
observe the kicking child to classify the video as breakdancing. The
perturbations in sample 3 (``mowing lawn'') do highlight a man pushing
a lawn mower. By removing intensity from the blue channels (instead
of the dominant green channels) modifications to the original image
are reduced. A subtle perturbation to sample 3 appears to have been
the blurring of the mowing man's legs perhaps through targeted mixing
in the time dimension. As a result the man is seen in the perturbed
video slightly closer to being sliding. This may have influenced I3D's
classification of the perturbed sample 3 video as ``toboganing''.
For sample 4 there are three original dominant classes: ``tasting
beer'', ``drinking beer'' (the correct label in the kinetic dataset),
and ``bartending''. These three classes are of course related. Although
the lighting in the original video is dark on the non-essential background,
the perturbations further highlighted the important parts of the video,
as can be seen in the difference video frames. The perturbed video
is classified as ``slacklining'' (i.e., balancing on a tight rope)
and ``deadlifting'' either of which would be challenging for a human
observer to identify. Sample 5 was originally correctly classified
as ``javelin throw''. The perturbations leaved the images in the
video largely decolored (particularly the originally bright blue sky).
Although the essential aspects of the video are not modified from
the point of view of a human observer, the perturbed video was classified
by I3D as ``ice skating''. By decoloring the images, the last frames
of the video are slightly closer to presenting two people scatting
on an ice ring.

\begin{table}
\resizebox{\columnwidth}{!}{%
\centering

\begin{tabular}{llllll} \toprule   & {} & \multicolumn{2}{l}{original} & \multicolumn{2}{l}{perturbated} \\   & {} &                             class &   score &                      class &   score \\ sample & {} &                                   &         &                            &         \\ \midrule 1 & {} &                      scuba diving &  0.8832 &             smoking hookah &   0.995 \\   & {} &                        snorkeling &   0.114 &                    smoking &   0.004 \\   & {} &                     cleaning pool &  0.0026 &               scuba diving &  0.0008 \\   & {} &                springboard diving &  0.0002 &  swimming butterfly stroke &  0.0001 \\   & {} &                 jumping into pool &  0.0001 &              surfing water &       0 \\   & {} &                                   &         &                            &         \\ 2 & {} &                      punching bag &  0.9183 &               breakdancing &  0.9656 \\   & {} &                      drop kicking &  0.0299 &              skateboarding &  0.0143 \\   & {} &                         side kick &  0.0226 &          jumpstyle dancing &  0.0053 \\   & {} &          punching person (boxing) &  0.0048 &                   krumping &  0.0039 \\   & {} &  exercising with an exercise ball &   0.003 &                tap dancing &  0.0023 \\   & {} &                                   &         &                            &         \\ 3 & {} &                       mowing lawn &  0.9681 &                tobogganing &  0.9982 \\   & {} &                   walking the dog &  0.0284 &                bobsledding &  0.0006 \\   & {} &                      training dog &  0.0034 &               training dog &  0.0004 \\   & {} &                    sweeping floor &       0 &            walking the dog &  0.0004 \\   & {} &                    blowing leaves &       0 &                mowing lawn &  0.0003 \\   & {} &                                   &         &                            &         \\ 4 & {} &                      tasting beer &   0.449 &                slacklining &  0.8925 \\   & {} &                     drinking beer &  0.2442 &                deadlifting &  0.0531 \\   & {} &                        bartending &  0.2127 &                headbanging &  0.0142 \\   & {} &                    opening bottle &  0.0458 &                   pull ups &  0.0057 \\   & {} &                          drinking &  0.0337 &            recording music &  0.0023 \\   & {} &                                   &         &                            &         \\ 5 & {} &                     javelin throw &  0.9893 &                ice skating &  0.9411 \\   & {} &                         long jump &  0.0056 &         playing ice hockey &  0.0188 \\   & {} &     catching or throwing softball &  0.0012 &             roller skating &  0.0178 \\   & {} &                   throwing discus &  0.0011 &             playing tennis &  0.0053 \\   & {} &                       triple jump &  0.0009 &                hockey stop &  0.0046 \\   & {} &                                   &         &                            &         \\ \bottomrule \end{tabular}

}

\caption{\label{tab:vid-origs-perts}Top 5 classes and associated I3D scores
for the original and perturbed images on the five video samples considered.}
\end{table}

\section{Discussion\label{sec:Discussion}}

\subsection{Contrast to related perturbative techniques\label{sub:Contrast-to-related}}

Among the several previously published perturbative methods for analyzing
deep learning models, the works by Fong and Vedaldi \cite{fong2017interpretable},
Wei et. al. \cite{wei2018sparse}, and Baluja and Fischer \cite{baluja2017adversarial}
share multiple similarities with the approach presented here. The
versatility of our perturbation procedure allowed us to treat both
image and video data. We are not aware of any previous perturbative
procedure applied to both types of data.

In their image analysis, Fong and Vedaldi \cite{fong2017interpretable}
learn a multiplicative mask for introducing blurring perturbations
on an original image. The optimum perturbations then both localize
the object in question and expose what the image classification model
(VGG19) considered relevant in reaching its decision. Additionally,
noise is also added to the images in order to avoid optimization artifacts.
In our case the perturbations are not limited to blurring but can
instead be more general. The convnets used here as perturbation generators
are able to create nonlinearly perturbed images and videos by passing
them through several stages of nonlinear transformations. This led
to the production of efficiently targeted sparse perturbations allowing
the resolution of details such as individual object features (see
the differences images in fig. \ref{fig:origs-perts-diffs}). Perhaps
due to the sparsity of the introduced modifications to the inputs,
we had no need for taking special precautions in order to avoid optimization
artifacts.

Wei et. al. directly modify the pixels in video frames to iteratively
reduce the original video's dominant class score. The added perturbations
are kept sparse by the addition of \emph{L1- }and \emph{L2}-norm loss
terms in the temporal and spatial dimensions, respectively. Additionally,
a mask is also used to explicitly prevent modifications to some frames.
In contrast, our video perturbations were generated by a convnet and
can thus acquire a larger degree of nonlinearity customized to a particular
sample. Moreover, we did not place any restrictions on what frames
could be perturbed.

Baluja and Fischer \cite{baluja2017adversarial} learned image perturbations
that confuse target models by training convnets on datasets containing
a number of image samples. In our case, perturbations are learned
from a single image or video sample. This choice has consequences
for the architectures used for generating the perturbations. Our perturbation
generator $P$ is intended to sequentially modify the input image
or video while keeping its dimensions unmodified. Thus, all convolutional
layers have a single input and a single output with the same dimensions
as the input image or video. Essentially, a slightly modified version
of the original image or video flows through the layers of $P$. This
was considered important in keeping the perturbed image or video close
to the input. Learning perturbations on a single sample has advantages
and disadvantages. On the one hand, the perturbations learned by our
method are specialized to the image or video in question. We are thus
able to produce perturbations with rich detail on each sample, as
shown above. On the other hand, having more data to learn perturbations
from, Baluja and Fischer were able to create perturbations that forced
target models to misclassify into arbitrarily chosen classes. Their
approach is complementary to ours.

\subsection{Object or action detection and explanation of model outcomes\label{sub:detection-explanation}}

The identification of portions of images or video frames that are
found to be strongly correlated with the top scores of a classification
model can be leveraged as a form of object or action detection \cite{selvaraju2017grad,chollet2017deep,wei2018sparse}.
In our case, the extracted sparse image or video differences (see
figs. \ref{fig:origs-perts-diffs} and \ref{fig:videos}) indicate
the locations of the most important features in the inputs. In contrast
with other localization techniques \cite{simonyan2013deep,springenberg2014striving,zeiler2014visualizing,zhang2016top,ribeiro2016should,selvaraju2017grad,fong2017interpretable},
the highlighted features in the image or video differences are generally
given here with a high level of detail and preserve the color information.

Another important aspect of the extracted perturbations is the insights
they provide about what the models consider important in reaching
a classification decision. Due to their increasing widespread application,
machine learning models in general and deep learning models in particular
are required to show transparency in their decision process \cite{albrecht2016gdpr}.
Additionally, understanding model outcomes is crucial for detecting
and correcting possible errors or for enhancing model performance
\cite{gunning2017explainable,chollet2017deep}. The high resolution
of the extracted perturbations shown above allow appreciating, in
detail, what features the models have considered important. For instance,
the eyes, nose, mouth, and even the red tongue of the dog in image
sample 1 (top of fig. \ref{fig:origs-perts-diffs}) are clearly delineated
in the corresponding differences image. To different degrees, it is
possible to understand the image and video perturbations in sec. \ref{sec:Experiments}
in intuitive terms. At the same time, the applied perturbations also
exposed flaws in the studied models, as discussed next.

\subsection{Enhancing model generalization through adversarial data augmentation\label{sub:augmentation}}

We discuss here an augmentation procedure similar to previously proposed
schemes involving retraining on adversarial examples \cite{szegedy2013intriguing,goodfellow6572explaining,madry2017towards}
but adapted to the perturbation method described in the present work.
As discussed in sec. \ref{sub:detection-explanation}, the studied
image and video classification models consider relevant features in
reaching their decisions. On the other hand, slight perturbations
can cause drastic misclassification (see figs. \ref{fig:origs-perts-diffs}
and \ref{fig:videos} and tables \ref{tab:img-origs-perts} and \ref{tab:vid-origs-perts}).
The root of this problem may lie in the weight that the models assign
to the various detected features. From a human's point of view, the
main features in the perturbed samples were still similar to the ones
in the original images or videos. Moreover, any inserted features
belonging to a different class in the perturbed samples were not important
enough to change the sample class from a human's perspective. This
situation can be mitigated by ``telling'' the model about their
misclassification through re-training \cite{szegedy2013intriguing,goodfellow6572explaining,madry2017towards}.
The retraining would consist on a semi-supervised process. In this
case, one would obtain an initial set of sparse perturbations (as
done here for images and videos) for each sample in a training set
and fine-tune the model with the correct labels (i.e., the original
sample labels) for the perturbed samples. For fine-tunning purposes,
the dataset does not have to be as large as the one used to originally
train the model (e.g., such as the imagenet dataset \cite{russakovsky2015imagenet}
for VGG19), although it is indeed possible to retrain on the entire
original dataset. Multiple perturbed instances for the same sample
can be obtained by simply running several times the perturbation algorithm.
It may also be possible to further augment the data by using other
sparse perturbations \cite{baluja2017adversarial,wei2018sparse}.
Once retrained, the model can be re-evaluated through a new round
of perturbations. At this point it is expected that less sparse perturbations
will be needed in order to induce the retrained model to misclassify.
The process can be repeated as many times as needed until the loss
no longer converges under a threshold value during the perturbation
procedure. For instance, the loss in our image and video experiments
was generally reduced to values under 0.2, but in some situations
(e.g., when attempting to change a dog image into a ``sea lion'')
the loss could not be made less than 1. Failure to drive down the
loss in eq. (\ref{eq:loss_min_top5}) can be an indication that the
class can not be modified without severely perturbing the original
sample. Human supervision after each model retraining iteration is
required for assessing whether or not perturbed samples still belong
to the original class.

\subsection{Image or Video compression and approximate reconstruction\label{sub:compression}}

The perturbations of images and videos in sec. (\ref{sec:Experiments})
were found to be sparse but at the same time detailed in the features
they represent (see the image or video differences in figs. (\ref{fig:origs-perts-diffs})
and (\ref{fig:videos})). We propose leveraging this concentrated
form of information into a compression scheme. To obtain a compressed
representation of an image or video first obtain a perturbed sample
through the process described in sec. (\ref{sec:Learned-perturbations})
and then calculate the difference $\textbf{{\ensuremath{\boldsymbol{\delta}}}}^{(i)}$
between the original and perturbed samples, 
\[
\textbf{{\ensuremath{\boldsymbol{\delta}}}}^{(i)}=\textbf{X}^{(i)}-P^{(i)}(\mathbf{X}^{(i)}),
\]
where $i$ is a sample index, $\mathbf{X}^{(i)}$ is an image or video
sample, and $P^{(i)}$ is a perturbation generator learned for $\mathbf{X}^{(i)}$.
Most of the elements of array $\textbf{\ensuremath{\boldsymbol{\delta}}}^{(i)}$
are expected to be approximately zero, as seen above. Thus, it is
possible to obtain a sparse array from $\textbf{\ensuremath{\boldsymbol{\delta}}}^{(i)}$
by thresholding its values. A more robust procedure for obtaining
a sparse representation of $\textbf{\ensuremath{\boldsymbol{\delta}}}^{(i)}$
is to apply to it a wavelet transform and threshold the result to
obtain a set of sparse coefficients $\boldsymbol{\omega}^{(i)}$,
i.e., 
\[
\boldsymbol{\omega}^{(i)}=T\left(W\left(\textbf{\ensuremath{\boldsymbol{\delta}}}^{(i)}\right)\right),
\]
where $W$ and $T$ denote a wavelet transform and thresholding operators,
respectively. The sparse coefficients $\boldsymbol{\omega}^{(i)}$
contain only essential information about sample $\textbf{X}^{(i)}$
from the point of view of the classification model $M$. Irrelevant
contextual information is removed from $\boldsymbol{\omega}^{(i)}$.

Approximate reconstruction could be achieved through training a neural
network $R$ as described next. The sparse coefficients $\boldsymbol{\omega}^{(i)}$
are first transformed to an image- or video-like array $\boldsymbol{\gamma}^{(i)}$
through an inverse wavelet transform 
\[
\boldsymbol{\gamma}^{(i)}=W^{-1}\left(\boldsymbol{\omega}^{(i)}\right).
\]
During training, the inputs of $R$ are given by the set of sparse
samples $\boldsymbol{\gamma}$ and its labels are the set of original
samples $\mathbf{X}$. The optimum reconstruction model $R$ is the
result of the following minimization

\[
\underset{R}{\text{min}}\left(\textbf{X}-R\left(\boldsymbol{\gamma}\right)\right).
\]
Model $R$'s architecture may consist of a set of deconvolutional
layers \cite{zeiler2014visualizing} for upsampling the input sparse
image (i.e., hence inducing the addition of context information to
it) followed by convolutional and maxpooling layers for downsampling
back to the original image or video dimensions. A trained model $R$
can then be deployed for reconstructing approximate image or video
samples $\boldsymbol{\chi}^{(i)}$ from compressed coefficients $\boldsymbol{\omega}^{(i)}$
by
\[
\boldsymbol{\chi}^{(i)}=R\left(W^{-1}\left(\boldsymbol{\omega}^{(i)}\right)\right).
\]

Lacking exact contextual information, the reconstructed images or
videos $\boldsymbol{\chi}^{(i)}$ would likely be far from an accurate
reconstruction of the original samples $\textbf{X}^{(i)}$. The value
of this scheme lies on the capability of transmitting a minimum compressed
amount of relevant information given by the coefficients $\boldsymbol{\omega}^{(i)}$.
In the case of videos it may be advantageous to add a term to the
loss for penalizing pixel value changes from frame to frame, effectively
inducing the generation of a continuous background across frames.
In some circumstances (e.g., the transmission of a movie under low-bandwidth
conditions) it may be advantageous to overfit model $R$ during training,
effectively inducing $R$ to ``memorize'' context information in
the training set (a desirable situation assuming that only images
or videos in the training set are to be transmitted).

\section{Conclusions\label{sec:Conclusions}}

We implemented a method for learning adversarial perturbations on
images and videos with respect to their classification by deep convolutional
network models. The sparse, but at the same time detailed, perturbations
found allow a form of object or action recognition. The perturbations
also provide insights into what features the models considered important
when reaching their classification decisions. The sparse adversarial
perturbations successfully confused the models into misclassifying
while still belonging to the original class from a human's point of
view. This may be leveraged for a proposed training procedure based
on adversarial data augmentation. The sparsity and high detail of
the learned perturbations may also be leveraged into a form of image
or video compression and approximate reconstruction.

\section*{Acknowledgments}

RRC acknowledges the support of the NSF (grant number CHE-1763198).
HR acknowledges the support of the DOE (grant number DE-FG02-02ER15344).
RRC also acknowledges the support of NVIDIA Corporation for the donation
of the Titan X GPU used for this research.

\bibliographystyle{unsrt}
\bibliography{bibliography}

\end{document}